\documentclass{article}
\usepackage{spconf,amsmath,graphicx,hyperref,booktabs,amssymb,xcolor,amsmath,bm,float,tabularx}
\usepackage{caption}
\usepackage{subcaption}

\title{SAIP: A Plug-and-Play Scale-adaptive Module in Diffusion-based Inverse Problems}
\name{Lingyu Wang\thanks{This work has been submitted to the IEEE for possible publication. 
Copyright may be transferred without notice, after which this version may no longer be accessible.}, Xiangming Meng$^*$\thanks{$^*$Corresponding author}}
\address{Zhejiang University, ZJU-UIUC Institute, Haining,China}
\begin{document}
\topmargin=0mm
\ninept
\maketitle
\begin{abstract}
Solving inverse problems with diffusion models has shown promise in tasks such as image restoration. A common approach is to formulate the problem in a Bayesian framework and sample from the posterior by combining the prior score with the likelihood score. Since the likelihood term is often intractable, estimators like DPS, DMPS, and $\pi$GDM are widely adopted. However, these methods rely on a fixed, manually tuned scale to balance prior and likelihood contributions. Such a static design is suboptimal, as the ideal balance varies across timesteps and tasks, limiting performance and generalization.
To address this issue, we propose SAIP, a plug-and-play module that adaptively refines the scale at each timestep without retraining or altering the diffusion backbone. SAIP integrates seamlessly into existing samplers and consistently improves reconstruction quality across diverse image restoration tasks, including challenging scenarios.
\end{abstract}
\begin{keywords}
Inverse problems; diffusion models; optimal scale; image restoration
\end{keywords}
\section{Introduction}
\label{sec:intro}

Image restoration is a typical inverse problem that aims to recover a clean, high-quality image \(x \in \mathbb{R}^{N \times 1}\) from a degraded observation \(y \in \mathbb{R}^{M \times 1}\). Depending on the degradation process, typical tasks include denoising~\cite{DBLP:journals/mmas/BuadesCM05}, deblurring~\cite{kawar2022denoising}, and inpainting~\cite{bertalmio2000image}, all of which can be cast as a linear inverse problem:
\begin{equation}
y = Ax + n,
\end{equation}
where \(A \in \mathbb{R}^{M \times N}\) is a linear forward operator and \(n \sim \mathcal{N}(0, \sigma^2 I)\) denotes Gaussian noise. Due to system limitations and environmental disturbances, the measurement \(y\) is often incomplete, noisy, or otherwise degraded. The resulting ill-posedness generally implies non-uniqueness of the solution \(x\), making restoration fundamentally challenging. To resolve this ambiguity and select a plausible solution, one must incorporate prior knowledge or regularization about the target signal. From a Bayesian perspective~\cite{stuart2010inverse}, this amounts to performing posterior inference \(p(x\mid y)\) via Bayes’ rule,
\begin{equation}
p(x\mid y) \propto p(x)\, p(y\mid x),
\label{eq:bayesrule}
\end{equation}
where \(p(x)\) and \(p(y\mid x)\) are the prior and the likelihood, respectively. The prior \(p(x)\) encapsulates assumptions about natural images and thus plays a central role in inverse problems. Classical choices include hand-crafted, analytical priors (e.g., sparsity~\cite{DBLP:journals/spm/CandesW08}, low-rank structure~\cite{fazel2008compressed}, and total variation~\cite{DBLP:journals/spm/CandesW08}) as well as architecture-induced priors such as the Deep Image Prior~\cite{ulyanov2018deep}. However, these priors often fall short of capturing the full complexity of natural images. Recently, generative models, especially diffusion models~\cite{nichol2021improved,DBLP:journals/corr/abs-2011-13456}, have emerged as powerful data-driven priors for inverse problems~\cite{DBLP:conf/iclr/ChungKMKY23,kawar2022denoising,DBLP:conf/acml/MengK24,song2023pseudoinverse,DBLP:journals/corr/abs-2410-09945,DBLP:journals/corr/abs-2501-18913,chen2025solving,uehara2025inference}. Following Bayes’ rule~\eqref{eq:bayesrule}, posterior inference with diffusion models requires the posterior score:
\begin{equation}
\nabla_{x_t} \log p(x_t|y) = \nabla_{x_t} \log p(x_t) + \lambda \nabla_{x_t} \log p(y|x_t),
\label{eq:estimatedlikelihood}
\end{equation}
where \(x_t\) is a noisy latent in the diffusion process and \(\lambda\) is a scaling factor. The prior score \(\nabla_{x_t} \log p(x_t)\) is provided by the pretrained diffusion model, whereas the likelihood score \(\nabla_{x_t} \log p(y\mid x_t)\) is generally intractable\cite{kawar2022denoising,janati2025mixture,DBLP:conf/acml/MengK24,DBLP:conf/iclr/ChungKMKY23}. A number of approximations have been proposed, including Diffusion Posterior Sampling (DPS)~\cite{DBLP:conf/iclr/ChungKMKY23}, Pseudoinverse-Guided Diffusion Models (\(\pi\)GDM)~\cite{song2023pseudoinverse}, and Diffusion Model Posterior Sampling (DMPS)~\cite{DBLP:conf/acml/MengK24}.

\begin{figure*}[t]
\centering
\includegraphics[width=0.95\linewidth]{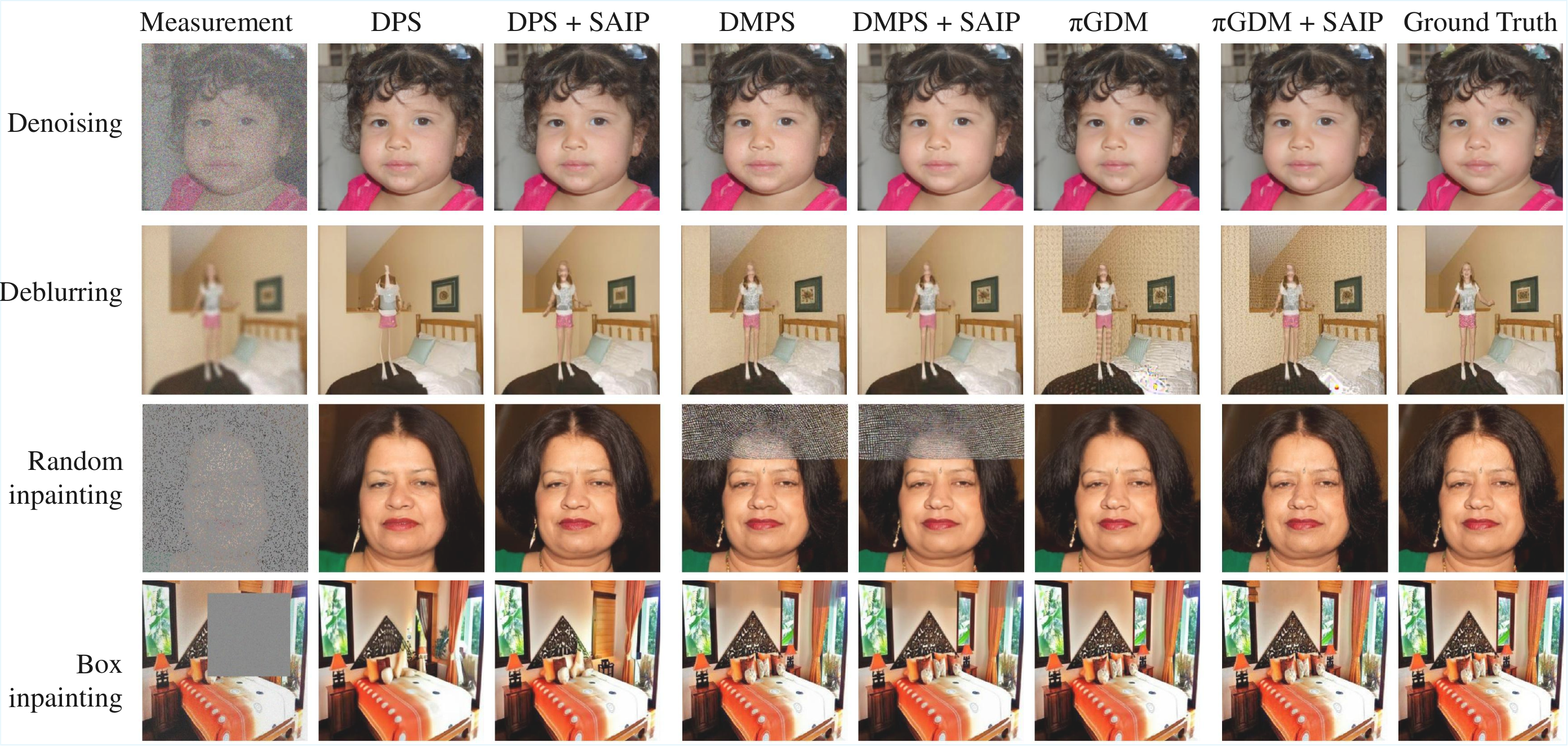}
\caption{Representative results. Rows 1 and 3: FFHQ at \(256 \times 256\); Rows 2 and 4: LSUN-Bedroom at \(256 \times 256\).}
\label{fig:fig1}
\end{figure*}

Despite their success, existing methods typically rely on a fixed, manually tuned \(\lambda\) to balance the prior and  likelihood contributions. This practice is tedious and often suboptimal. As observed in DMPS~\cite{DBLP:conf/acml/MengK24}, the final reconstruction quality is highly sensitive to \(\lambda\): even small deviations can markedly degrade sampling performance or lead to instability. Moreover, a single, manually chosen coefficient lacks robustness across noise levels and timesteps, necessitating extensive per-setting tuning. This not only reduces efficiency but also exposes \(\lambda\) selection as a key bottleneck that limits the stability and generalization of current diffusion-based inverse solvers.

To address this, inspired by the self-adaptive guidance in CFG-Zero*~\cite{DBLP:journals/corr/abs-2503-18886} for conditional generation, we extend the idea to inverse problems and propose a self-adaptive scaling module (\textbf{SAIP}). SAIP dynamically estimates \(\lambda\) during inference, allowing the guidance strength at each timestep to adapt to task and noise specific characteristics, thereby combining prior and likelihood in a more robust manner. SAIP is plug-and-play: it integrates into existing diffusion-based inverse problem solvers without retraining or modifying the diffusion backbone. Comparisons between the baseline and its "SAIP +" version are shown in Fig.~\ref{fig:fig1}.

% Inspired by the self-adaptive guidance in CFG-Zero*\cite{DBLP:journals/corr/abs-2503-18886} for conditional generation, we extend it to inverse problems and propose a self-adaptive scaling module (\textbf{SAIP}). SAIP dynamically estimates \(\lambda\) during inference, adapting guidance strength at each timestep to task and noise characteristics, thus combining prior and likelihood more robustly. It is plug-and-play, integrating into existing diffusion-based solvers without retraining or modifying the backbone. Comparisons between baseline and SAIP version are shown in Fig.\ref{fig:fig1}.

The main {contributions} are summarized as follows:

\noindent
\textbf{(1)} We extend the idea of adaptive scaling to inverse problems and introduce SAIP, a plug-and-play module that dynamically estimates the scale during inference. To the best of our knowledge, this is the first work to introduce adaptive scaling into diffusion-based inverse problem solvers. SAIP also can be directly applied to a wide range of methods that combine estimated likelihood and prior scores, such as DiffStateGrad~\cite{zirvi2024diffusion}, without retraining. \\
\textbf{(2)} We integrate SAIP into several representative diffusion-based solvers and evaluate it on classical image restoration tasks. Experiments span both standard settings and more challenging scenarios with severe degradation, high noise and latent domain. The results demonstrate consistent improvements over nearly all baselines, while incurring only a minor additional inference cost.

\section{METHOD}

To address the fixed scale challenge in diffusion-based inverse problem solvers, we propose a plug-and-play module called \textbf{SAIP},
% (\underline{S}cale-\underline{a}daptive Module in Diffusion-based \underline{I}nverse \underline{P}roblems)
 which dynamically adjusts the balance between the prior and likelihood scores during posterior sampling.

Our design is inspired by the self-adaptive guidance introduced in CFG-Zero*~\cite{DBLP:journals/corr/abs-2503-18886} for classifier-free guidance~\cite{DBLP:journals/corr/abs-2207-12598}:
\begin{equation}
\nabla_{x_t} \log \tilde{p}(x_t|y) = \omega \nabla_{x_t} \log p(x_t|y) + s(1 - \omega) \nabla_{x_t} \log p(x_t),
\label{eq:cfgzero}
\end{equation}
where $\omega$ is the guidance strength and $s$ is a learnable scaling factor.

To adapt this idea to inverse problems, we replace the posterior score in right side of Eq.\eqref{eq:cfgzero} with the sum of the likelihood score and the prior score.
\begin{equation}
\begin{aligned}
    \nabla_{x_t} \log \tilde{p}(x_t | y)
    % &= \omega \Big[ \nabla_{x_t} \log p(y|x_t) 
    %    + \nabla_{x_t} \log p(x_t) \Big] \\
    % &\quad + s(1 - \omega) \nabla_{x_t} \log p(x_t) \\
    &= \big[ s(1 - \omega) + \omega \big] 
       \nabla_{x_t} \log p(x_t) \\
    &\quad + \omega \nabla_{x_t} \log p(y | x_t).
\end{aligned}
\label{eq:SAIP_form}
\end{equation}
Then, we reorganize the equation by factoring out the prior score term with a coefficient of 1 from the posterior score expression.
\begin{equation}
\begin{aligned}
    \nabla_{x_t} \log \tilde{p}(x_t | y)
    &= \underbrace{\nabla_{x_t} \log p(x_t)
       + \omega \nabla_{x_t} \log p(y | x_t)}_{\text{as same as Eq.~\eqref{eq:estimatedlikelihood} when $\omega = \lambda$}} \\
&\quad + \underbrace{\bm{\left[ (s - 1)(1 - \omega) \right] \nabla_{x_t} \log p(x_t)}}_{\textbf{Additional part}} ,
\end{aligned}
\label{eq:SAIP_form_reorganized}
\end{equation}
Thus, compared to Eq.~\eqref{eq:estimatedlikelihood}, the SAIP formulation essentially adds an adaptive offset $(s - 1)(1 - \omega)$ to the prior score, making it possible to refine the balance between the two components adaptively. \textbf{In particular, when setting $s = 1$, Eq.~\eqref{eq:SAIP_form_reorganized} reduces to the original formulation of the baseline method.}

The remaining problem is to calculating $s$, we minimize the gap between the estimated posterior score and the ground-truth posterior score:
\begin{equation}
    \mathcal{L}(s) = \left\| \nabla_{x_t} \log \tilde{p}_\theta(x_t|y) - \nabla_{x_t} \log p^*(x_t|y) \right\|^2,
    \label{eq:loss}
\end{equation}
where the first term is the estimated posterior score modulated by SAIP, as defined in Eq.~\eqref{eq:SAIP_form}, and $p^*(x_t|y)$ denotes the true posterior. Since $p^*$ is intractable, We replace the posterior score in Eq.~\eqref{eq:SAIP_form_reorganized} and introduce a new parameter $\omega^\star = \omega - 1$, which allows us to rewrite the loss as
\begin{equation}
\mathcal{L}(s) =
\left\|
\begin{aligned}
& \nabla_{x_t} \log \tilde{p}_\theta(x_t|y) - \nabla_{x_t} \log p^*(x_t|y) \\
& \quad + \omega^\star \left( -s \nabla_{x_t} \log p(x_t) + \nabla_{x_t} \log p(y|x_t) \right)
\end{aligned}
\right\|^2.
\label{eq:lossstar}
\end{equation}

Applying the triangle inequality, we obtain
\begin{align}
\mathcal{L}(s) \leq & \;
\left\| \nabla_{x_t} \log p_\theta(x_t\mid y) \right\|^2 \nonumber + \left\| \nabla_{x_t} \log p^*(x_t\mid y) \right\|^2 \nonumber \\
& + \left\| \omega^\star \left( -s \nabla_{x_t} \log p(x_t) 
   + \nabla_{x_t} \log p(y \mid x_t) \right) \right\|^2,
\end{align}
which shows that minimizing $\mathcal{L}(s)$ is equivalent to minimizing the upper-bound
\begin{equation}
\mathcal{L}_{\text{upper}} =
\left\|
-s \nabla_{x_t} \log p(x_t) + \nabla_{x_t} \log p(y|x_t)
\right\|^2.
\label{eq:loss_upper}
\end{equation}

To minimize the $\mathcal{L}_{\text{upper}}$ , we take the derivative with respect to $s$ and set it to zero:
\begin{align}
\frac{d}{ds} \mathcal{L}_{\text{upper}}
&= 2s \left\| \nabla_{x_t} \log p(x_t) \right\|^2 \nonumber \\
&\quad - 2 \left\langle \nabla_{x_t} \log p(x_t), 
\nabla_{x_t} \log p(y \mid x_t) \right\rangle = 0.
\label{derivation}
\end{align}

Finally, we obtain the closed-form optimal value of $s$ .

\begin{equation}
    s = \frac{ \langle \nabla_{x_t} \log p_\theta(x_t|y), \nabla_{x_t} \log p_\theta(x_t) \rangle }{ \| \nabla_{x_t} \log p_\theta(x_t) \|^2 },
    \label{eq:s_star}
\end{equation}
which can be computed in closed form. 
It is worth noting that \textbf{the additional cost of SAIP is very low}, and adding SAIP does not increase the additional reasoning cost. More detailed discussions on time and memory consumption can be found in Section~\ref{sec:computational}.

\section{Experiments}
In this section, we conduct experiments on several inverse problems to evaluate the effectiveness of the proposed \textbf{SAIP} module. The code is available at: \url{https://github.com/seulaugues/SAIPcode}.

\begin{figure*}[t]
\centering
\includegraphics[width=0.48\linewidth]{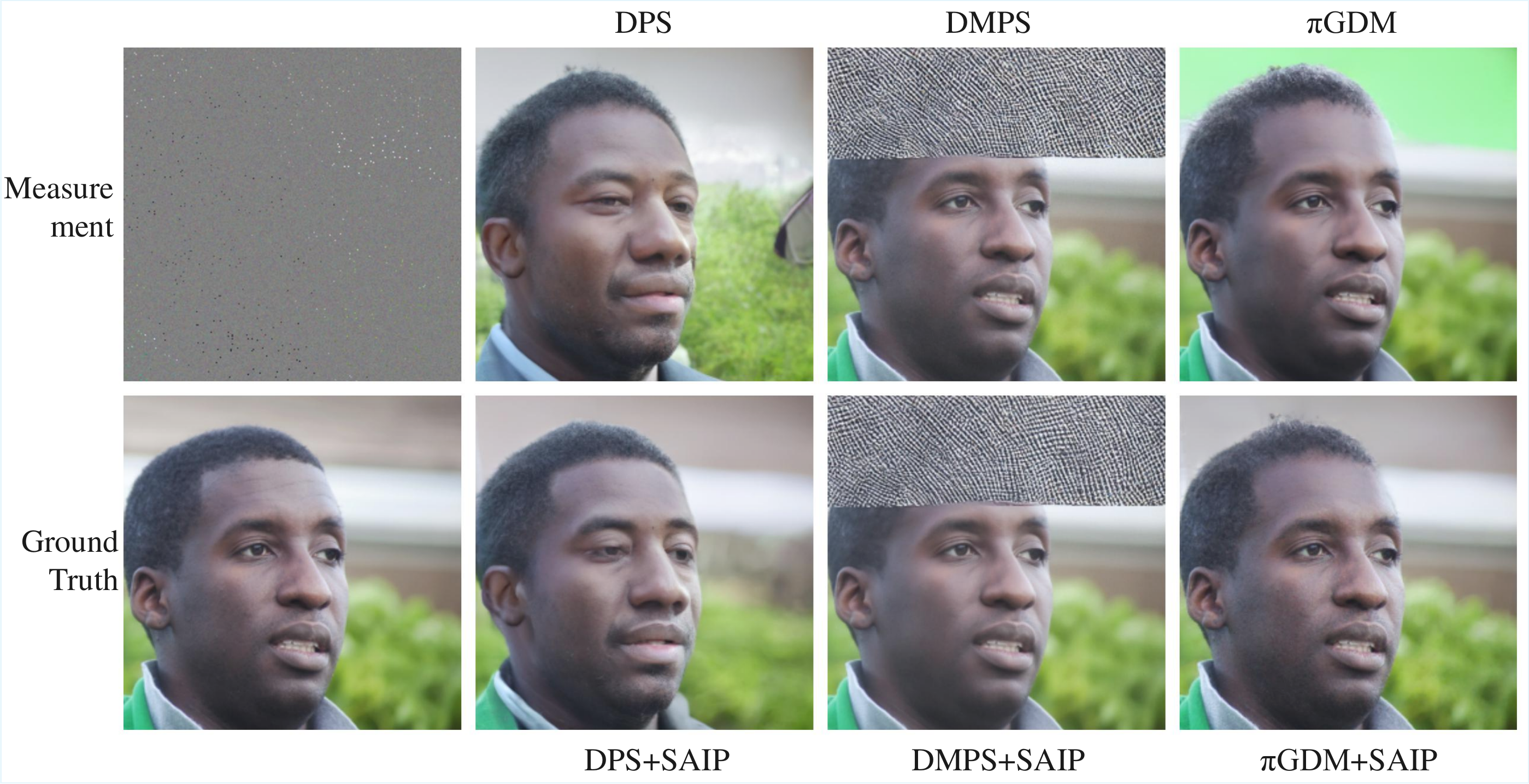}
\includegraphics[width=0.48\linewidth]{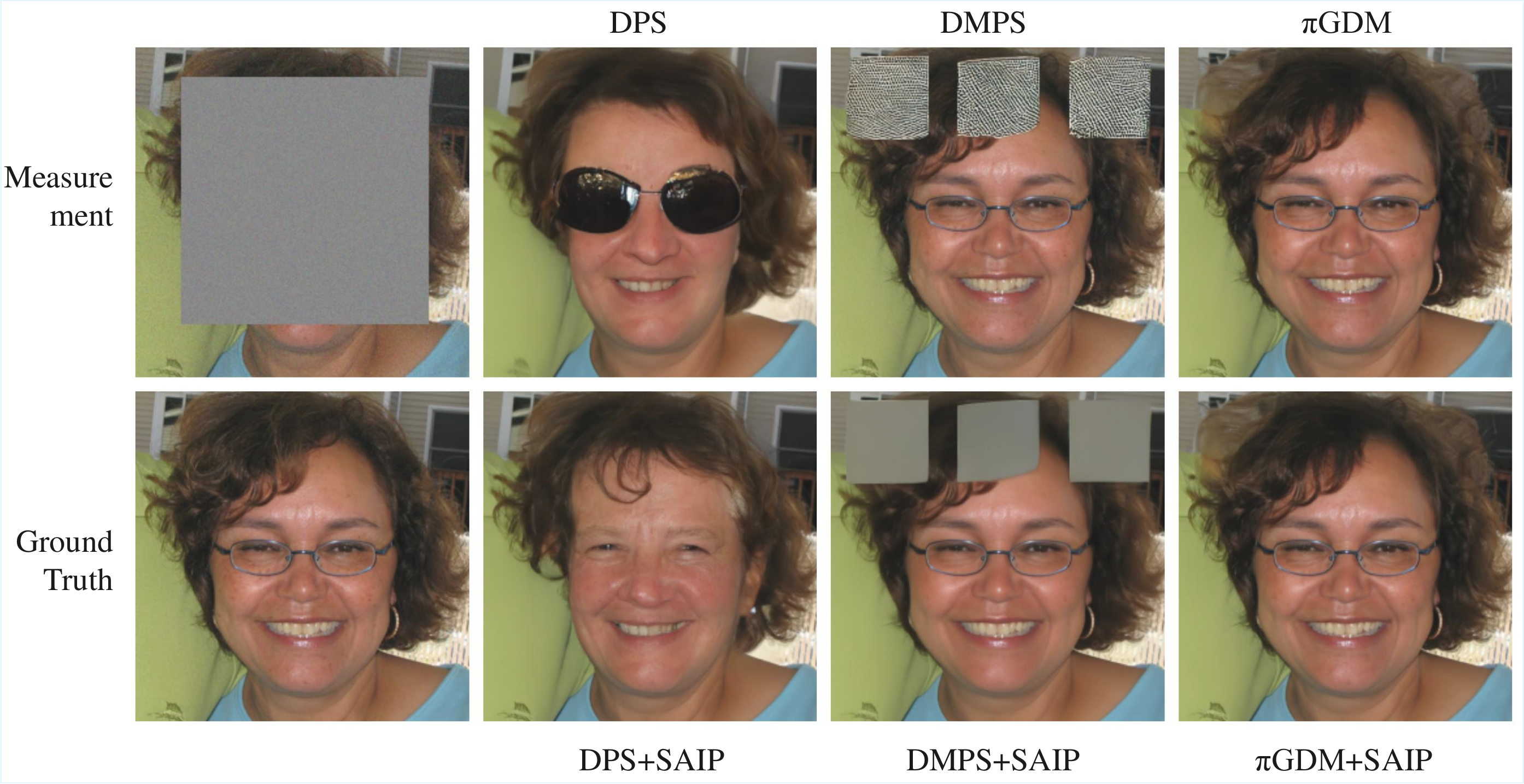}
\centerline{(a) Random inpainting 99\% \quad (b) Box inpainting 191$\times$191}\medskip

\includegraphics[width=0.48\linewidth]{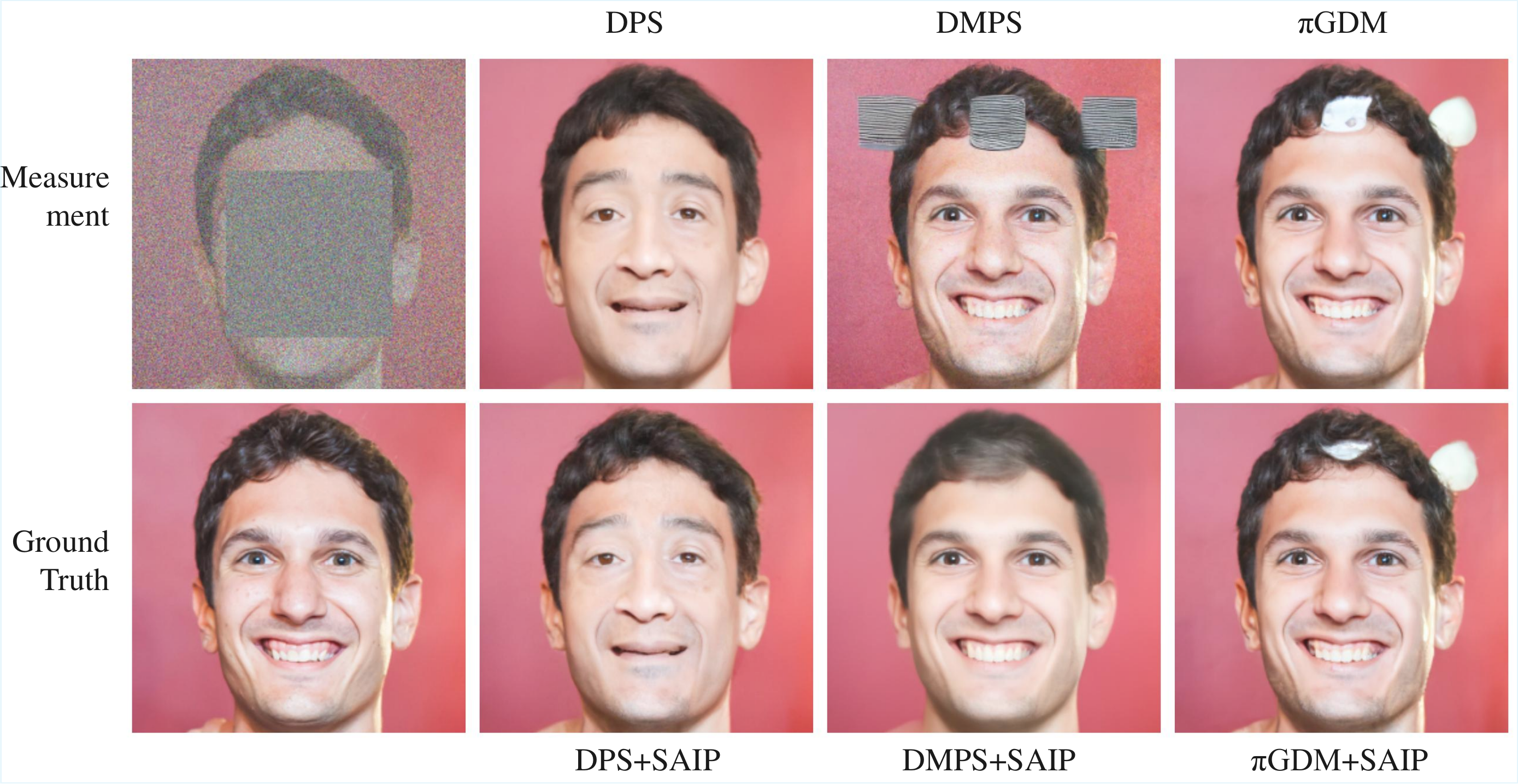}
\includegraphics[width=0.48\linewidth]{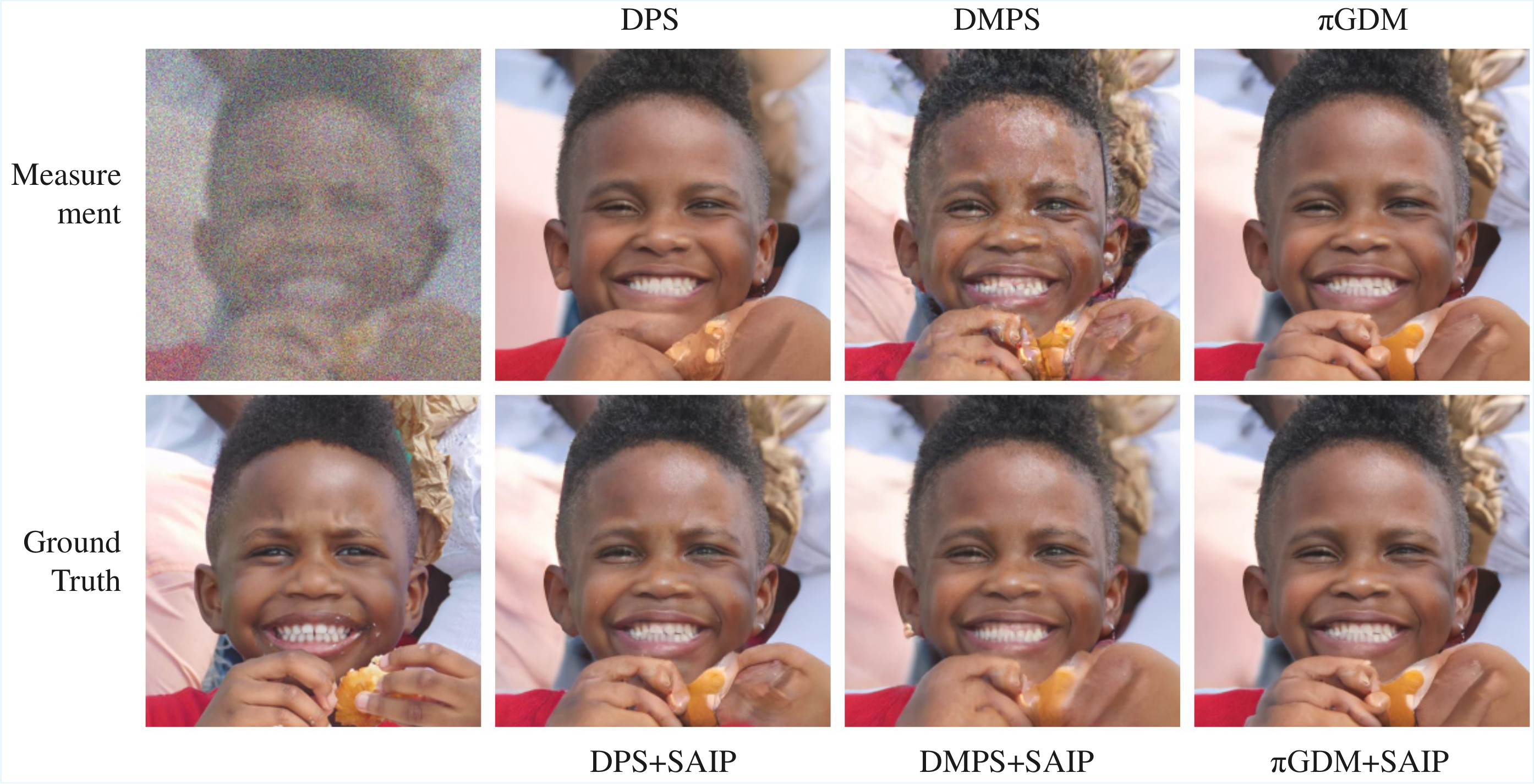}
\centerline{(c) Box inpainting $\sigma = 0.5$ \quad (d) Uniform deblur $\sigma = 0.5$}\medskip
\caption{Challenging results for different inverse tasks.}
\label{fig:fig2}
\end{figure*}

\subsection{Experimental Setup}

\subsubsection{Tasks.}

\textbf{Standard:} 
(a) Uniform deblurring~\cite{kawar2022denoising} with a $9\times9$ uniform kernel; 
(b) Denoising~\cite{DBLP:journals/mmas/BuadesCM05} with Gaussian noise ($\sigma=0.5$); 
(c) Random inpainting~\cite{bertalmio2000image} with 90\% missing pixels; 
(d) Box inpainting~\cite{bertalmio2000image} with a $128 \times 128$ square mask. 
A mild Gaussian noise ($\sigma=0.05$) is added in all cases except denoising.\\
\textbf{Challenging:} 
(a) Intensified degradation: $\sigma=0.9$ denoising, $191 \times 191$ box inpainting, and 99\% random inpainting; 
(b) High-level noise injection: Gaussian noise increased from $\sigma=0.05$ to $0.5$ for deblurring/inpainting.
(c) Latent domain and nonlinear transfer: Based on DiffStateGrad~\cite{zirvi2024diffusion} + Resample (main sampler)~\cite{Song2023SolvingIP}, $128 \times 128$ box inpainting, 70\% random inpainting, and $2\times$ high dynamic range.

\subsubsection{Other Details.}
Datasets are FFHQ~\cite{DBLP:conf/cvpr/KarrasLA19} and LSUN-bedroom~\cite{DBLP:journals/corr/YuZSSX15} ($256\times256$), which we conducted experiments on. We adopt pretrained ADM diffusion models\cite{choi2021ilvr} (FFHQ from Google Drive, LSUN-bedroom from OpenAI GitHub). Baselines include DPS, DMPS, and $\pi$GDM, with "SAIP +" versions denoted DPS+SAIP, DMPS+SAIP, and $\pi$GDM+SAIP. Performance is evaluated using PSNR\cite{hore2010image}, SSIM~\cite{DBLP:journals/tip/WangBSS04}, and LPIPS~\cite{DBLP:conf/cvpr/ZhangIESW18}, and all experiments are run on a single NVIDIA RTX 4090.

\begin{table*}[t]
\centering
\ninept
\caption{Quantitative comparison on different tasks under standard settings on 1k images of the FFHQ $256 \times 256$ dataset. We report only mean values for clarity, as SAIP shows the variance comparable to or slightly smaller than that of the baselines. Best results are in \textbf{bold}.}
\label{table:ffhqtable}
\begin{tabularx}{\textwidth}{l *{12}{>{\centering\arraybackslash}X}}
\toprule
& \multicolumn{3}{c}{\textbf{Denoising}} 
& \multicolumn{3}{c}{\textbf{Deblurring}} 
& \multicolumn{3}{c}{\textbf{Random inpainting}} 
& \multicolumn{3}{c}{\textbf{Box inpainting}} \\
\cmidrule(lr){2-4} \cmidrule(lr){5-7} \cmidrule(lr){8-10} \cmidrule(lr){11-13}
\textbf{Method} & PSNR & LPIPS & SSIM
& PSNR & LPIPS & SSIM
& PSNR & LPIPS & SSIM
& PSNR & LPIPS & SSIM \\
\midrule
DPS
& 27.47 & 0.2405 & 0.7737
& 25.36 & 0.2576 & 0.7184
& 24.32 & 0.2915 & 0.7067
& \textbf{23.31} & 0.2148 & 0.7993 \\
+ SAIP
& \textbf{27.50} & \textbf{0.2365} & \textbf{0.7796}
& \textbf{27.18} & \textbf{0.2028} & \textbf{0.7863}
& \textbf{26.83} & \textbf{0.1967} & \textbf{0.8090}
& 22.83 & \textbf{0.1705} & \textbf{0.8303} \\
\midrule
DMPS
& 27.76 & 0.2473 & 0.7628
& 27.25 & 0.2241 & 0.7681
& 16.47 & 0.3356 & 0.6581
& 20.15 & 0.1766 & 0.8641 \\
+ SAIP
& \textbf{28.15} & \textbf{0.2098} & \textbf{0.8034}
& \textbf{27.47} & \textbf{0.2194} & \textbf{0.7877}
& \textbf{17.27} & \textbf{0.3299} & \textbf{0.6720}
& \textbf{24.86} & \textbf{0.1737} & \textbf{0.8866} \\
\midrule
$\pi$GDM
& 27.28 & 0.2311 & 0.7814
& 27.10 & \textbf{0.2225} & 0.7700
& 30.43 & 0.1393 & 0.8919
& 28.51 & \textbf{0.1044} & \textbf{0.9220} \\
+ SAIP
& \textbf{27.30} & \textbf{0.2301} & \textbf{0.7817}
& \textbf{27.12} & \textbf{0.2226} & \textbf{0.7708}
& \textbf{30.97} & \textbf{0.1260} & \textbf{0.8996}
& \textbf{28.81} & 0.1091 & 0.9210 \\
\bottomrule
\end{tabularx}
\end{table*}

\begin{table*}[t]
\centering
\ninept
\caption{Quantitative comparison on FFHQ $256 \times 256$ validation dataset under high-level noise challenging tasks (type b): \textbf{Denoising} with $\sigma = 0.9$; \textbf{other tasks} with Gaussian noise $\sigma = 0.5$. Best results are in \textbf{bold}.}
\label{table:ffhqchallenging}
\begin{tabularx}{\textwidth}{l *{12}{>{\centering\arraybackslash}X}}
\toprule
& \multicolumn{3}{c}{\textbf{Denoising}} 
& \multicolumn{3}{c}{\textbf{Deblurring}} 
& \multicolumn{3}{c}{\textbf{Random inpainting}} 
& \multicolumn{3}{c}{\textbf{Box inpainting}} \\
\cmidrule(lr){2-4} \cmidrule(lr){5-7} \cmidrule(lr){8-10} \cmidrule(lr){11-13}
\textbf{Method} 
& PSNR & LPIPS & SSIM
& PSNR & LPIPS & SSIM
& PSNR & LPIPS & SSIM
& PSNR & LPIPS & SSIM \\
\midrule
DPS
& 25.00 & 0.3018 & 0.6809
& 22.28 & 0.3359 & 0.6089
& 19.41 & 0.4194 & 0.5215
& 21.42 & 0.3159 & 0.6570 \\
+ SAIP
& \textbf{25.17} & \textbf{0.2913} & \textbf{0.6963}
& \textbf{24.55} & \textbf{0.2839} & \textbf{0.6793}
& \textbf{22.52} & \textbf{0.3381} & \textbf{0.6167}
& \textbf{21.52} & \textbf{0.2849} & \textbf{0.6920} \\
\midrule
DMPS
& 25.32 & 0.3208 & 0.6529
& 21.94 & 0.3725 & 0.5329
& 16.10 & 0.4521 & 0.5149
& 20.11 & 0.3341 & 0.6688 \\
+ SAIP
& \textbf{25.82} & \textbf{0.2669} & \textbf{0.7185}
& \textbf{23.04} & \textbf{0.3175} & \textbf{0.6268}
& \textbf{16.88} & \textbf{0.3984} & \textbf{0.6398}
& \textbf{22.04} & \textbf{0.3188} & \textbf{0.7002} \\
\midrule
$\pi$GDM
& 24.61 & 0.2985 & 0.6827
& 21.29 & 0.3592 & 0.5544
& 24.92 & 0.2703 & 0.7251
& \textbf{25.16} & 0.2476 & \textbf{0.7579} \\
+ SAIP
& \textbf{24.72} & \textbf{0.2927} & \textbf{0.6893}
& \textbf{21.79} & \textbf{0.3410} & \textbf{0.5803}
& \textbf{24.95} & \textbf{0.2692} & \textbf{0.7254}
& 25.00 & \textbf{0.2469} & 0.7569 \\
\bottomrule
\end{tabularx}
\end{table*}

\subsection{Results}

\subsubsection{Qualitative results.}

Fig.~\ref{fig:fig2} presents representative qualitative results under challenging conditions. For subplot (a), the integration of DPS effectively removes the artifact on the right side and yields a subject appearance more consistent with the ground truth. In the case of $\pi$GDM, it successfully eliminates the large green background region, resulting in a more natural subject. For subplot (b), DPS removes the eyeglasses and introduces bangs compared to the baseline, while DMPS produces a smoother transition around the mask box on the subject’s head. For subplot (c), incorporating SAIP into DPS leads to sharper details and more pronounced facial wrinkles, DMPS directly removes the box artifact, and $\pi$GDM reduces the residual box region in the upper part of the image. For subplot (d), SAIP enhances DPS by generating clearer textures and distinct boundaries on the subject’s fingers, while in DMPS, SAIP suppresses sharp artifacts and yields a smoother overall appearance. 

Across all examples, SAIP consistently improves reconstruction quality over baselines, producing better structural fidelity, coherent textures, and perceptual realism, even under extreme degradations. Under both standard and high-level noise settings, integrating SAIP consistently improves the best performing baseline methods across denoising, deblurring, and inpainting tasks.

\subsubsection{Quantitative results.}

\begin{table}[htb]
\centering
\ninept
\caption{Part of supplementary results on LSUN-bedroom $256 \times 256$ dataset.}
\begin{tabular}{lccc}
\toprule
Task & PSNR $\uparrow$ & LPIPS $\downarrow$ & SSIM $\uparrow$ \\
\midrule
Denoise (DMPS)        & 26.94 & 0.3052 & 0.7177 \\
\quad + SAIP             & \textbf{27.77} & \textbf{0.2234} & \textbf{0.8106} \\
\midrule
Deblur ($\pi$GDM)   & 24.83 & 0.2854 & 0.6900 \\
\quad + SAIP             & \textbf{25.98} & \textbf{0.2483} & \textbf{0.7541} \\
\midrule
Random Inpaint ($\pi$GDM) & 29.64 & 0.1465 & 0.8974 \\
\quad + SAIP             & \textbf{32.64} & \textbf{0.0992} & \textbf{0.9325} \\
\midrule
Box Inpaint ($\pi$GDM)    & 26.64 & 0.1081 & 0.9273 \\
\quad + SAIP             & \textbf{28.49} & \textbf{0.0909} & \textbf{0.9332} \\
\bottomrule
\label{table:lsun_results}
\end{tabular}
\end{table}

\begin{table}[H]
\centering
\ninept
\caption{Quantitative comparison on FFHQ $256 \times 256$ validation dataset under intensified degradation tasks (type a). }
\begin{tabular}{lccc}
\toprule
Task & PSNR $\uparrow$ & LPIPS $\downarrow$ & SSIM $\uparrow$ \\
\midrule
Denoise (DMPS)        & 25.32 & 0.3208 & 0.6529 \\
\quad + SAIP             & \textbf{25.82} & \textbf{0.2669} & \textbf{0.7185} \\
\midrule
Random Inpaint (DPS)   & 18.54 & 0.4295 & 0.4968 \\
\quad + SAIP             & \textbf{20.19} & \textbf{0.3467} & \textbf{0.5843} \\
\midrule
Box Inpaint (DMPS)     & 16.37 & 0.2669 & 0.7539 \\
\quad + SAIP             & \textbf{19.76} & \textbf{0.2455} & \textbf{0.8133} \\
\bottomrule
\label{table:challenging_results_typea}
\end{tabular}
\end{table}

We present a quantitative comparison based on standard and challenging evaluation settings. Table~\ref{table:ffhqtable} reports the reconstruction performance of different algorithms on the FFHQ $256 \times 256$ dataset under standard settings, while supplementary experiments on the LSUN-bedroom dataset (Table~\ref{table:lsun_results}) further confirm that SAIP consistently enhances the performance of all baseline methods. Note that for consistency, all reported numbers are mean values, as the variance is slightly lower than that of the baselines. Due to space constraints, only a subset of the LSUN results is reported.

In challenging settings, SAIP consistently improves upon baseline methods, as summarized in Table~\ref{table:challenging_results_typea}, Table~\ref{table:ffhqchallenging}, and Table~\ref{table:Diffstategrad_results}. Even under conditions with severe degradations, strong noise interference, or operation in the latent domain, SAIP demonstrates clear gains over the original solvers. These results indicate that SAIP is not confined to particular types of diffusion-based methods, but can be effectively applied across diverse restoration scenarios, spanning both pixel- and latent-domain formulations. The observed improvements further highlight its applicability, adaptability to different tasks, and robustness under both linear and nonlinear degradation conditions.

\begin{table}[H]
\centering
\ninept
\caption{Challenging tasks (type c) on DiffStateGrad + Resample (origin, FFHQ $256 \times 256$ dataset). }
\begin{tabular}{lccc}
\toprule
Task & PSNR $\uparrow$ & LPIPS $\downarrow$ & SSIM $\uparrow$ \\
\midrule
Random Inpaint (Origin)        & 31.20 & 0.1621 & 0.8925 \\
\quad + SAIP              & \textbf{31.36} & \textbf{0.1587} & \textbf{0.8966} \\
\midrule
Box Inpaint (Origin)           & \textbf{21.07} & 0.2501 & 0.7828 \\
\quad + SAIP              & 21.06 & \textbf{0.2477} & \textbf{0.7839} \\
\midrule
HDR (Origin)           & 24.60 & 0.2840 & 0.7706 \\
\quad + SAIP              & \textbf{24.71} & \textbf{0.2824} & \textbf{0.7736} \\
\bottomrule
\label{table:Diffstategrad_results}
\end{tabular}
\end{table}

\subsubsection{Computational Overhead.}
\label{sec:computational}
As shown in Table.~\ref{table:runtimeandmemory}, all experiments are conducted on a single NVIDIA RTX 4090 for denoising on FFHQ, and incorporating SAIP increases per-image inference time by 0.5s and memory usage by 3MB, which is modest and acceptable in practice.

\begin{table}[H]
\centering
\ninept
\caption{Runtime (s) and memory consumption (MB) for different algorithms on the FFHQ dataset for the denoising task. }
\label{table:runtimeandmemory}
\begin{tabular}{lccc}
\toprule
Metric & DPS / +SAIP & DMPS / +SAIP & $\pi$GDM / +SAIP \\
\midrule
Time    & 90.01 / 90.32 & 32.67 / 33.00 & 73.04 / 73.65 \\
Memory  & 5449 / 5453   & 4492 / 4496   & 2782 / 2785 \\
\bottomrule
\end{tabular}
\end{table}

\subsection{S-Curve Analysis}

\begin{figure}[t]
\centering
% 第一排两个
\begin{subfigure}{0.45\linewidth}
    \includegraphics[width=\linewidth]{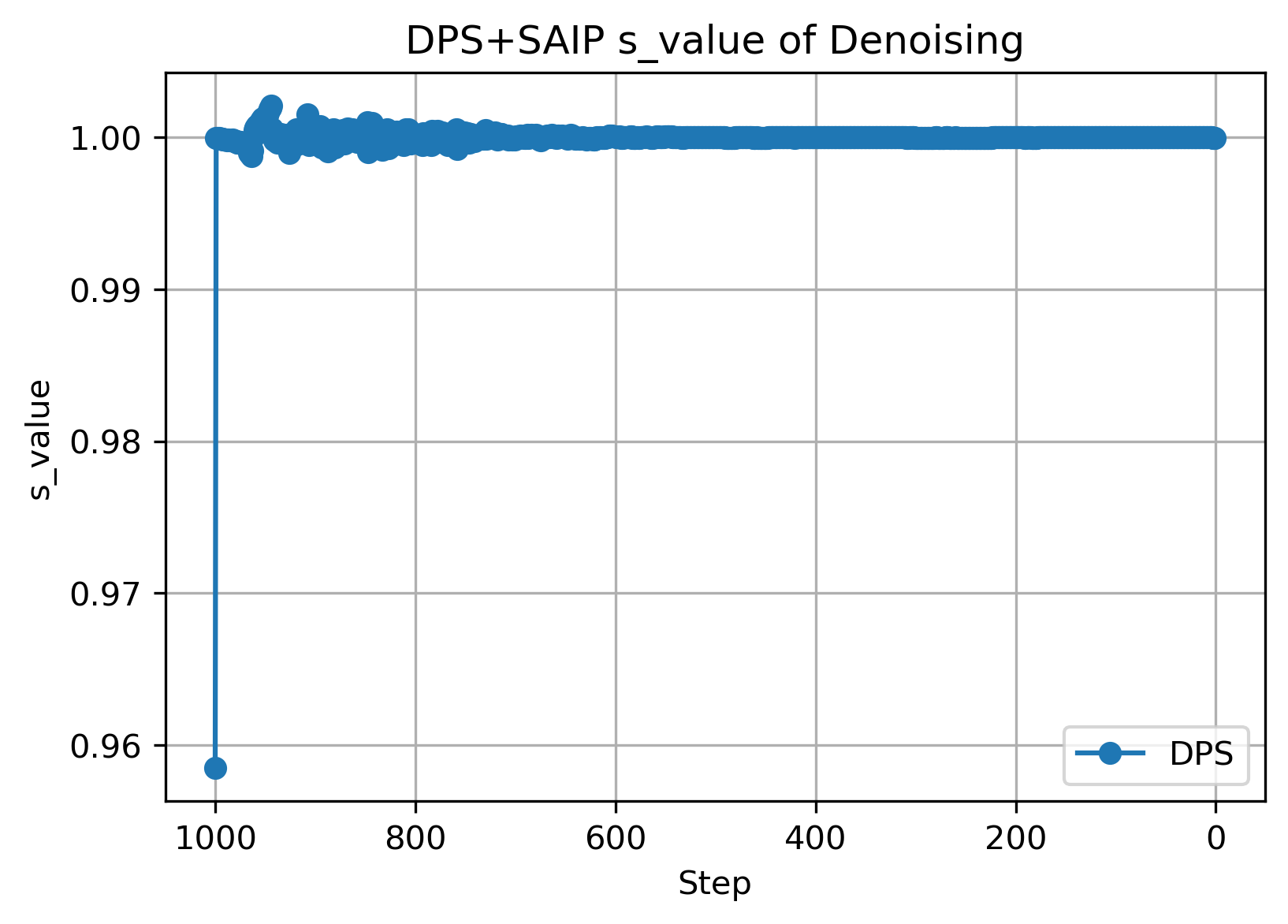}
    \caption{DPS + SAIP (denoising)}
\end{subfigure}
\begin{subfigure}{0.45\linewidth}
    \includegraphics[width=\linewidth]{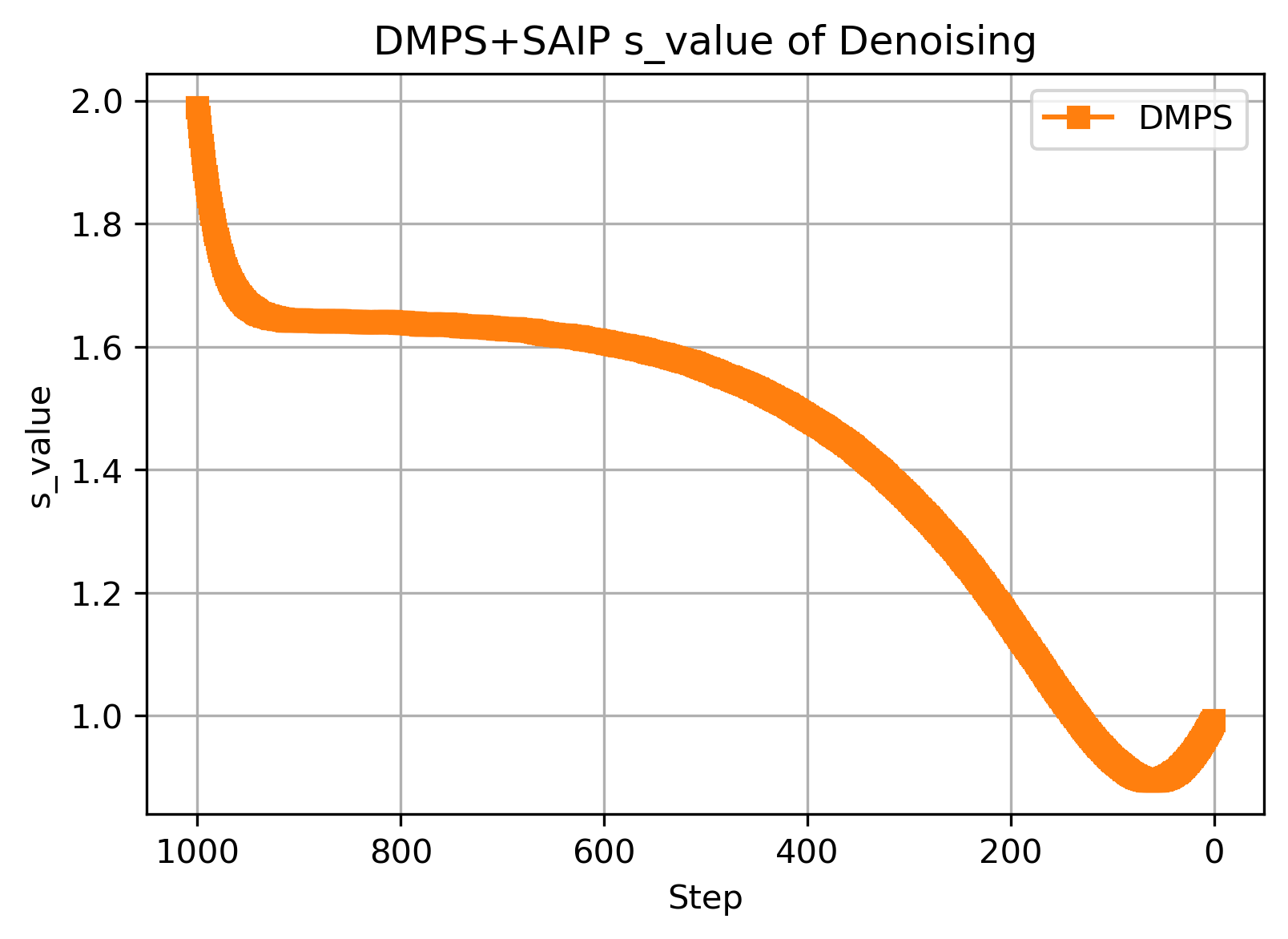}
    \caption{DMPS + SAIP (denoising)}
\end{subfigure}

% 第二排一个（换行）
\begin{subfigure}{0.45\linewidth}
    \includegraphics[width=\linewidth]{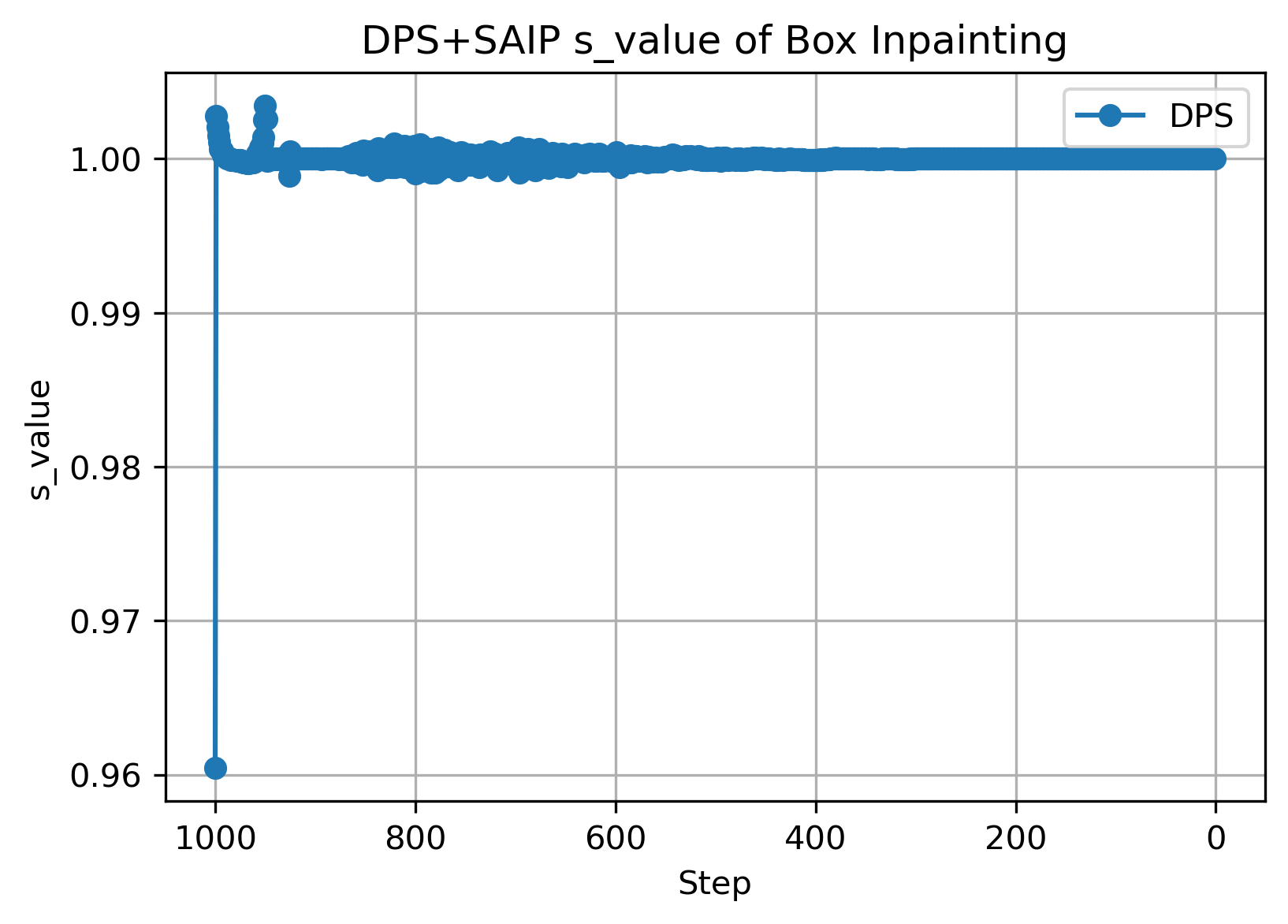}
    \caption{DPS + SAIP (box inpainting)}
\end{subfigure}

\caption{Scale $s$ for SAIP across different methods and tasks. Figures (a) and (b) present successful cases, (c) shows a failure case.}
\label{fig:s-curve}
\end{figure}

Fig.~\ref{fig:s-curve} shows the adaptive scale $s$ along the reverse diffusion trajectory (t = 1000 to 0) under the same DDPM~\cite{DBLP:journals/corr/abs-2006-11239} schedule. Subplots (a) and (b) show successful cases, while (c) shows a failure case. The denoising task is used as a example, since $s$ trends similarly across tasks, and the $\pi$GDM curve closely follows DPS and is omitted.

Across all cases, $s$ initially deviates from 1 and gradually returns as the process progresses, reflecting early-stage inaccuracies of the likelihood score estimator~\cite{DBLP:conf/iclr/ChungKMKY23,DBLP:conf/acml/MengK24,song2023pseudoinverse}. By balancing prior and likelihood, $s$ mitigates the bias and eventually converges toward 1.

The key difference between success and failure lies in early stage and mid stage behavior. In successful cases (Fig.~\ref{fig:s-curve}a,b), $s$ corrects the likelihood bias immediately and stabilizes, preventing error accumulation and enabling accurate reconstructions. In contrast, the failure case (Fig.~\ref{fig:s-curve}c) also exhibits an initial oscillation, but its duration is short, after which $s$ remains with only small fluctuations around 1. It only resumes noticeable activity in the middle stage, by which time the trajectory has already deviated from the optimal path. This delayed response limits SAIP’s corrective power and ultimately leads to a failed reconstruction. These results indicate that prompt early adjustment of $s$ is critical for SAIP’s success.

% The key difference between success and failure lies in the early- and mid-stage behavior of $s$. In successful cases (Fig.~\ref{fig:s-curve}a,b), $s$ starts to take effect immediately: it undergoes a noticeable oscillation at the beginning, which mitigates the inaccuracy of the likelihood score prediction in the early stage. As the likelihood score becomes more accurate over time, the oscillations of $s$ gradually decay toward 1, thereby approaching the baseline behavior. In contrast, the failure case (Fig.~\ref{fig:s-curve}c) also exhibits an initial oscillation, but its duration is short, after which $s$ remains with only small fluctuations around 1. It only resumes noticeable activity in the middle stage, by which time the trajectory has already deviated from the optimal path. This delayed response limits SAIP’s corrective power and ultimately leads to a failed reconstruction. These results indicate that prompt early adjustment of $s$ is critical for SAIP’s success.

\section{Conclusion and Discussion}
\textbf{SAIP mechanism.} The core idea of SAIP is to compute a closed-form adaptive guidance scale $s$ by projecting the mismatch between the posterior score and the prior score onto the prior-score direction. This formulation provides three properties: 
(a) in the early stages of sampling, when the estimated likelihood score is often noisy, 
$s$ deviates from 1 to emphasize the contribution of the prior and prevent error accumulation; 
(b) as alignment improves during sampling, $s$ naturally converges toward 1, effectively recovering the behavior of the baseline method and ensuring stability; 
(c) in rare cases of score anti-alignment, $s$ may take transient negative values, which act as a corrective mechanism that stabilizes the trajectory.\\
\textbf{Conclusion.} We propose SAIP, a plug and play module that adaptively scales the balance between prior and (approximated) likelihood in diffusion based inverse problem solvers. In addition to integration with DPS, DMPS, and $\pi$GDM, SAIP can also be applied to other methods that combine prior and likelihood,such as DiffstateGrad, and it consistently improves reconstruction quality across challenging settings, while preserving computational efficiency. The results suggest that adaptive scaling has potential for improving diffusion-based solvers.

\vfill
\pagebreak

% \section{REFERENCES}
% \label{sec:refs}

% List and number all bibliographical references at the end of the
% paper. The references can be numbered in alphabetic order or in
% order of appearance in the document. When referring to them in
% the text, type the corresponding reference number in square
% brackets as shown at the end of this sentence \cite{C2}. An
% additional final page (the fifth page, in most cases) is
% allowed, but must contain only references to the prior
% literature.

% Please follow the IEEE Citation Guidelines, \url{https://ieee-dataport.org/sites/default/files/analysis/27/IEEE\%20Citation\%20Guidelines.pdf} for formatting of references.

% % References should be produced using the bibtex program from suitable
% % BiBTeX files (here: strings, refs, manuals). The IEEEbib.bst bibliography
% % style file from IEEE produces unsorted bibliography list.
% % -------------------------------------------------------------------------
\bibliographystyle{IEEEbib}
\bibliography{Template}

\end{document}